\documentclass[letterpaper, 10 pt, conference]{ieeeconf}  
\newcommand{\conj}[1]{\overline{#1}}

\IEEEoverridecommandlockouts                              

\usepackage{graphicx}
\usepackage{cite}
\usepackage{amsmath, multirow, booktabs, cleveref}

\title{\LARGE \bf
Body-Hand Modality Expertized Networks with Cross-attention for Fine-grained Skeleton Action Recognition
}

\author{Seungyeon Cho$^{1}$ and Tae-Kyun Kim$^{1}$
\thanks{$^{1}$School of Computing, KAIST, South Korea.}%
}

\begin{document}

\maketitle
\thispagestyle{empty}
\pagestyle{empty}

\begin{abstract}
Skeleton-based Human Action Recognition (HAR) is a vital technology in robotics and human–robot interaction. 
However, most existing methods concentrate primarily on full‐body movements and often overlook subtle hand motions that are critical for distinguishing fine-grained actions. 
Recent work leverages a unified graph representation that combines body, hand, and foot keypoints to capture detailed body dynamics.
Yet, these models often blur fine hand details due to the disparity between body and hand action characteristics and the loss of subtle features during the spatial-pooling.
In this paper, we propose BHaRNet (Body–Hand action Recognition Network), a novel framework that augments a typical body-expert model with a hand-expert model.
Our model jointly trains both streams with an ensemble loss that fosters cooperative specialization, functioning in a manner reminiscent of a Mixture-of-Experts (MoE).
Moreover, cross-attention is employed via an expertized branch method and a pooling-attention module to enable feature-level interactions and selectively fuse complementary information.
Inspired by MMNet, we also demonstrate the applicability of our approach to multi-modal tasks by leveraging RGB information, where body features guide RGB learning to capture richer contextual cues.
Experiments on large-scale benchmarks (NTU RGB+D 60, NTU RGB+D 120, PKU-MMD, and Northwestern-UCLA) demonstrate that BHaRNet achieves SOTA accuracies—improving from 86.4\% to 93.0\% in hand-intensive actions—while maintaining fewer GFLOPs and parameters than the relevant unified methods.

\end{abstract}
    
\section{INTRODUCTION}

\label{sec:intro}

Human Action Recognition (HAR) has garnered significant attention in robotics\cite{wen2023interactive, noisylabels}, especially in areas of human-robot interaction and collaboration. The goal of HAR is to automatically identify human actions from video data—a challenging task due to various factors such as complex temporal dependencies, intricate spatial relationships, background noise, and varying camera perspectives. To address these challenges, existing studies have explored diverse data modalities such as RGB video, depth, infrared, audio, and skeleton joints.

Skeleton-based approaches, particularly those using Graph Convolutional Networks (GCNs)\cite{yan2018stgcn, shi2019twoagcn} have shown great promise in efficiently modeling spatial-temporal dynamics by representing skeletal joints as graph nodes and their connections as edges. Recent architectures\cite{wang20233mformer, zhou2024blockgcn, lee2023hierarchically, chen2021ctrgcn, chi2022infogcn} have demonstrated strong performance in action recognition; however, most of these methods primarily focus on full-body motion analysis. This narrow focus limits their ability to capture the fine-grained hand movements essential for differentiating actions such as “Okay sign" or “Victory sign".

\begin{figure}[htbp]
    \centering
    \begin{minipage}[t]{0.48\columnwidth}
        \centering
        \includegraphics[width=\linewidth]{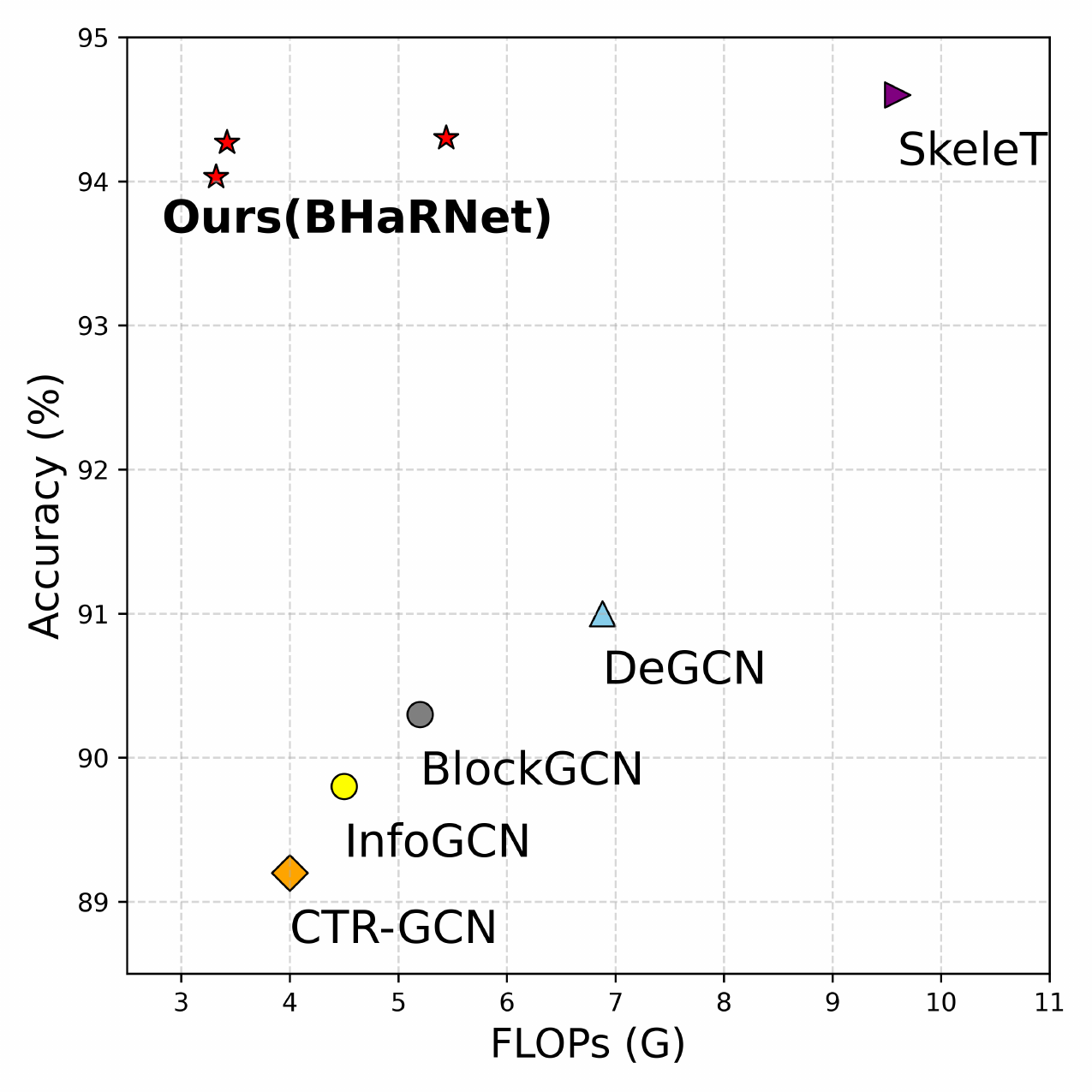} 
    \end{minipage}
    \hfill
    \begin{minipage}[t]{0.48\columnwidth}
        \centering
        \includegraphics[width=\linewidth]{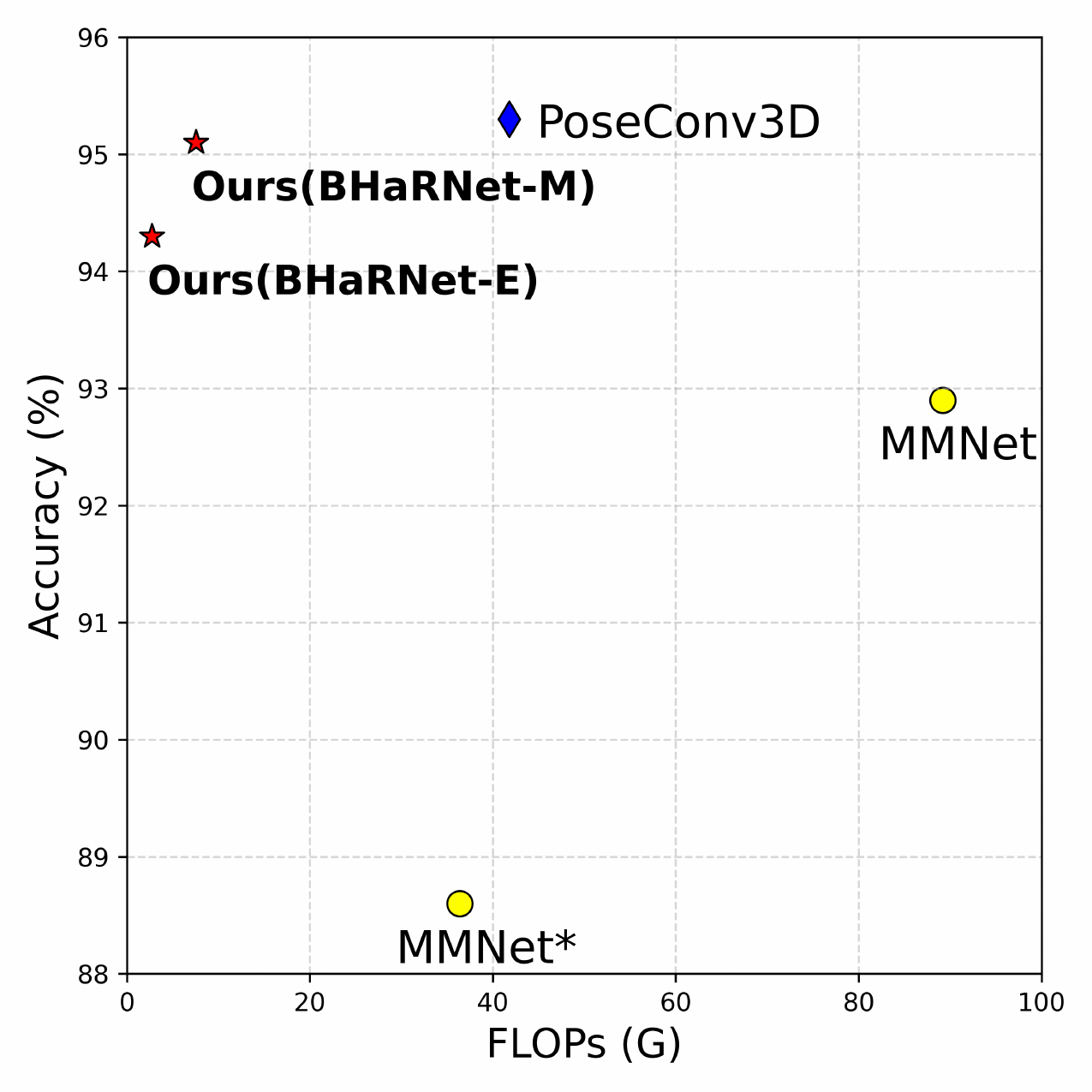} 
    \end{minipage}
    \caption{Performance vs. FLOPs in skeleton-based action recognition(left) and multi-modal action recognition(right). \textbf{Our models(red star)} shows competitive accuracy at lower GFLOPs compared to existing SOTA methods in both tasks. MMNet* is the multi-modal backbone of our model, with ResNet18.}
    \label{fig:two_plots}
\end{figure}

To address these limitations, recent research\cite{yang2024expressive} extends traditional skeleton-based representations by incorporating additional hand and foot keypoints into a unified graph. 
To manage the increased complexity of this enriched graph, a combined pooling strategy that integrates both node pooling and instance pooling is employed in this approach. 
While this combined pooling method in an largely-integrated graph significantly enhances efficiency and overall recognition performance, it also results in the loss of subtle hand details, making the hand features vulnerable to blur from larger, dynamic body movements.
Among the other related tasks, separating the body and hand into different graph streams is used in sign language recognition\cite{li2025unisign} and action forecasting methods\cite{yan2024forecasting, ding2024expforcasteai} (see Section II for more details).

In this work, we propose \textbf{BHaRNet} (\textbf{B}ody–\textbf{H}and \textbf{a}ction \textbf{R}ecognition \textbf{Net}work), a novel dual-stream model that augments a body-expert model with a hand-expert model.
In our design, the body model extracts features from body skeleton data to capture global pose dynamics, while a Hand model processes hand pose data to learn subtle hand articulations. 
The two stream models are jointly learnt via an ensemble loss as well as individual classification losses on its logits. The simple ensemble model at the score level, thanks to clear distinctiveness of two model output distributions, behaves as a mixture of experts where the modality-specific expertizes are maintained yet the two models cooperate in the intersection area. 

We also propose feature-level interaction via cross-attention as well as the above score-level fusion. By exchanging features at an intermediate stage—rather than just combining final logits—the body and hand streams can cooperatively interact with their expertizes.

Each modality is further refined in its dedicated expertized stream. Then, dividing interactive branch and expertized branch enables the two branches to exchange information without using residual connections—thereby preventing conflicts between expertized features and interactive features. Global average pooling is followed by an attention mechanism aggregating body and hand features. Integrating diversely-pooled interactive features, our approach results in a lightweight model with significantly reduced computational complexity.

We further equip each modality with an \emph{expertized branch model} to preserve subtle modality-specific cues and minimize information dilution. These branches operate in parallel with interactive (cross-attention) branches, thereby balancing the retention of crucial hand or body features with a controlled feature exchange. By sharing minimal overhead between the streams, our approach remains computationally efficient, particularly when partially pooling spatio-temporal dimensions before cross-attention.

Inspired by MMNet~\cite{bruce2022mmnet}, we leverage body skeleton features to guide the learning of RGB features. By treating body features as dynamic weights, our framework enables the RGB stream to focus on complementary contextual information, such as object interactions and environmental cues, enhancing overall recognition performance. 

Our contributions can be summarized as follows:\\
\noindent 1) We propose a novel cross-modal architecture that explicitly models body and hand as two complementary data modalities. It integrates detailed hand pose information with full-body poses in a cross-modal manner.\\
\noindent 2) The dual-modal framework is jointly trained via the ensemble loss as well as individual classification losses. The experiments show that the ensemble model exhibits two distinctive expertises, i.e. hand and body, yet improves over mixed classes.\\
\noindent 3) Our model dynamically balances body and hand contributions in the feature level using the cross attention, thereby yielding four branches, one expertized branch and one cooperative branch for each of hand or body. The pooling-based design helps obtain a lightweight model.\\
\noindent 4) We achieve significant improvements in accuracy and efficiency for skeleton-based action recognition, and further demonstrate robust performance in a multi-modal setting by integrating complementary RGB information. 

\section{RELATED WORK}

\label{sec:formatting}
\subsection{Skeleton-Based Action Recognition}
Skeleton-based action recognition has evolved significantly, moving from temporal modeling approaches such as RNNs, to more advanced graph-based methods like GCNs~\cite{yan2018stgcn, shi2019twoagcn}, CNNs\cite{duan2022poseconv}, and Graph-based Transformers~\cite{wang20233mformer, duan2023skeletr, do2025skateformer, pang2022igformer, abdelfattah2024maskclr}. RNNs effectively modeled sequential temporal dependencies but struggled with spatial relationships. GCNs addressed these challenges by representing human joints as graph nodes and their connections as edges, enabling simultaneous spatial and temporal dependency modeling. Graph-based Transformer frameworks further incorporate attention for long-range dependencies.

Recent GCN-based approach with SkeleT\cite{yang2024expressive} integrates body, hand, and foot keypoints into a unified graph to capture holistic human dynamics. Although this unified strategy yields strong performance in overall action recognition, it often results in feature dilution: the pooling operations required to handle the scale discrepancy can blur the subtle details of hand motions. This limitation is particularly evident in tasks that demand high precision in hand gesture recognition.

In contrast, our proposed BHaRNet decouples the processing of body and hand features into separate streams. In doing so, we maintain the integrity of fine-grained hand details while still capturing global body dynamics, addressing the key limitation observed in unified approaches (see ~\Cref{sec:experiments} for comparison of hand-intensive actions with the SkeleT model).

\begin{figure}[t]
    \centering
    \flushright
    \includegraphics[width=0.4\textwidth]{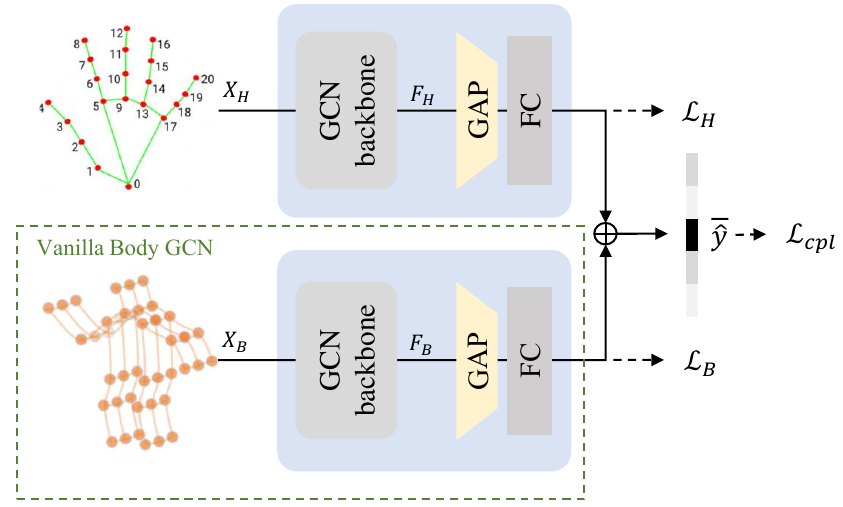}
    \caption{Overview of our proposed action recognition model. The model is a dual-stream network consists of Body and Hand expert models, utilizing GCN backbones and cross-attention for enhanced feature fusion.}
    \label{fig:overview}
\end{figure}

\subsection{Multi-Modal Action Recognition}
Multi-modal fusion has emerged as a promising strategy in action recognition, where modalities like RGB, depth, and skeleton data are jointly exploited for robust performance. 
For example, MMNet \cite{bruce2022mmnet} is noted for its lightweight design and strong robustness when combining a GCN backbone with an RGB backbone, effectively balancing skeletal and visual information. 
PoseConv3D\cite{duan2022poseconv} uses heatmap-based skeletal learning and bilateral interactions between RGB and skeletal streams, yielding effective multi-modal feature integration.
 A modular nature of our approach via MMNet facilitates additional
 modalities (e.g., RGB) without excessive overhead.

\begin{figure*}[t]
    \centering
    \includegraphics[width=0.8\textwidth]{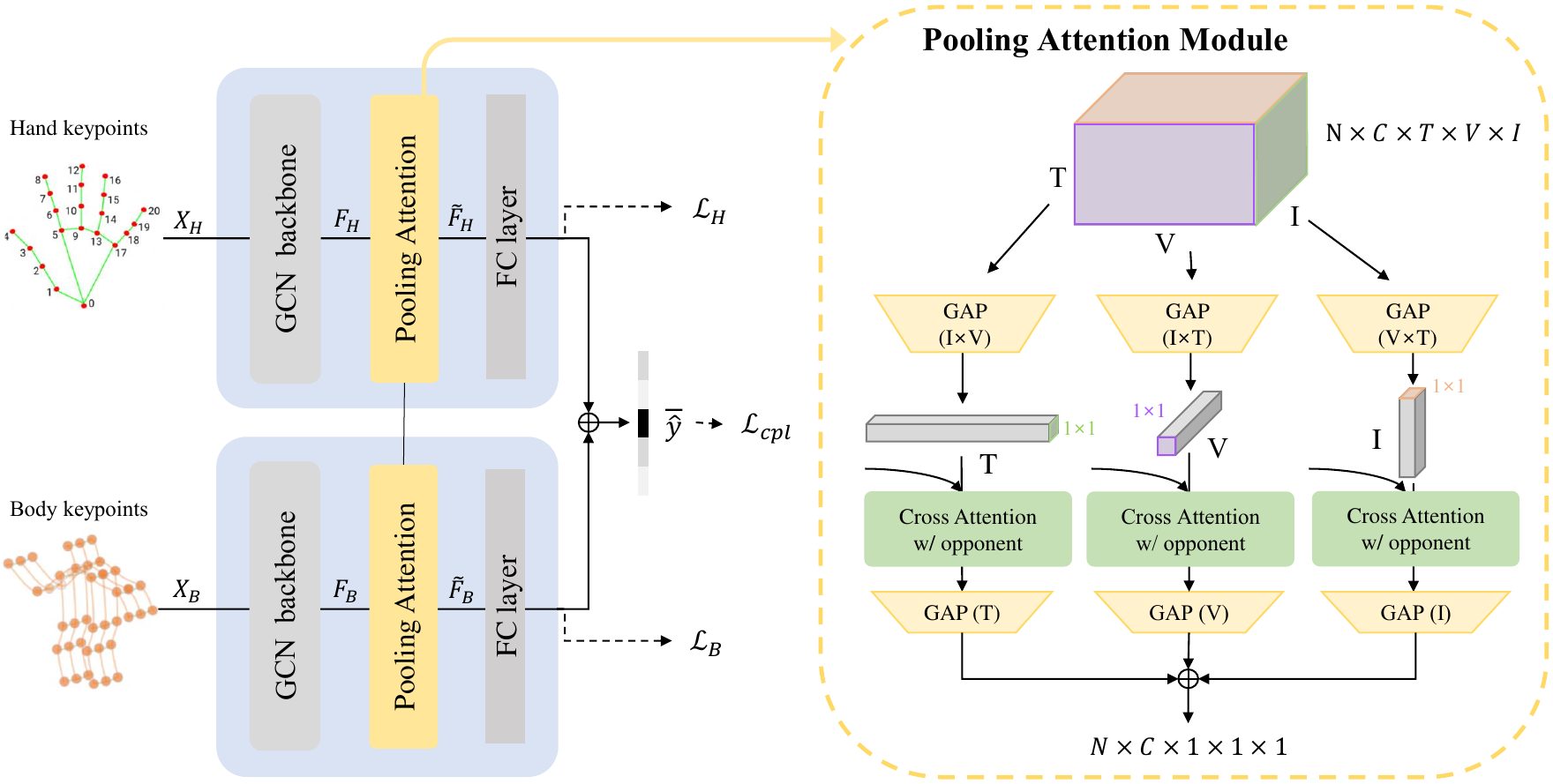}
    \caption{Shortcut of our Pooling Attention Module simplifying cross attention mechanism.}
    \label{fig:pooling}
\end{figure*}
\subsection{Body-Hand Keypoint Integration in Other Fields}
Beyond HAR, body–hand integration appears in several neighboring domains. 
For sign‑language recognition, UNI‑SIGN~\cite{li2025unisign} embeds body, hand, and face keypoints into a shared LLM‑based representation. Forecasting approaches~\cite{yan2024forecasting, ding2024expforcasteai} predict future body–hand coordination via specialized alignment modules, and REWIND~\cite{lee2025rewind} demonstrates the value of accurate body–hand keypoints for egocentric pose estimation.
Distinct from these methods, we preserve body and hand as separate experts and employ lightweight cross‑attention, avoiding the need for the shared embedding or the location precision. 

\subsection{Fine‑grained Hand Interaction}
Recent studies emphasize that preserving subtle finger articulations is crucial for understanding or synthesizing hand‑centric behavior.  
On the recognition side~\cite{shamil2024utility,kwon2021h2o,cho2023transformer}, accurate 3D or dual‑hand signals improved fine‑grained classification.  
Other studies~\cite{li2025latenthoi,woo2024hand} model detailed hand–object contacts, while two‑hand reconstruction approaches~\cite{lee2023im2hands,kim2024bitt} recover realistic geometry or texture from monocular input.  
These lines of research underline the importance of preserving fine‑grained hand cues.
\section{BHARNET}

\subsection{Overview}
In this section, we present \textbf{BHaRNet}, a dual-stream network designed to effectively capture both global body dynamics and fine-grained hand articulations. 
Our model treats the hand as an expert modality, processed independently ofhe body stream to better capture its unique characteristics.
The overview of our framework is illustrated in Figure~\ref{fig:overview}.

We first introduce the dual-stream architecture for body and hand, followed by our complementary loss formulation. 
We then describe a novel cross-attention mechanism that enables controlled feature interaction between the two streams. 
Finally, we extend our approach to integrate additional RGB data into a multi-modal setting.

Our architecture consists of separate GCN backbones: one specializing in full-body dynamics and another focused on intricate hand movements. Specifically, a hand GCN allows for specialized learning of finger movements, whereas a body GCN focuses on broader pose dynamics. By maintaining separate initial streams, our model effectively preserves modality-specific features.
\begin{equation} \label{eq:2}
    \mathbf{F}_{B}
    = \mathrm{GCN}_{B}(\mathbf{X}_{B}),\quad
    \mathbf{F}_{H}
    = \mathrm{GCN}_{H}(\mathbf{X}_{H})
\end{equation}
\subsection{Loss functions}
\label{subsec:loss}
A key element of BHaRNet is balancing the specialization of each modality with the overall synergy. We combine separate classification losses for body and hand with complementary loss that considers the fused prediction. Specifically, each modality produces a logit vector($\hat{y_B}, \hat{y_H}$) for its own classification task, leading to individual cross-entropy terms($\mathcal{L}_{B}, \mathcal{L}_{H}$) with the ground-truth label($l$). The fused logits, integrated by averaging logit vectors, are treated as a separate classifier with an additional loss term. Our loss implemented in joint training are:
\begin{gather}
    \mathcal{L}_{B} = \mathrm{CE}\bigl(\hat{y_B},\,l\bigr),\quad
    \mathcal{L}_{H} = \mathrm{CE}\bigl(\hat{y_H},\,l\bigr)\\
    \mathcal{L}_{\text{cpl}} = \mathrm{CE}\bigl(\conj{\hat{y}},\,l\bigr)\\
    \mathcal{L}_{\text{total}} = \mathcal\lambda_{B}\mathcal{L}_{B} \;+\;\mathcal\lambda_{H}\mathcal{L}_{H} \;+\;\lambda_{cpl}\,\mathcal{L}_{\text{cpl}}
\end{gather}
This design ensures the body stream retains its global pose knowledge, while the hand stream focuses on intricate gestures. The complementary loss also prevents the individual streams from being over-learned across all action classes, since the gradient stops if one of the streams reason enough on each action class. The logit distributions of the two models are clearly distinctive yielding the simple ensemble that effectively acts as a mixture-of-experts (MoE) mechanism. 

\begin{figure*}[htbp]
    \centering
    \begin{minipage}[t]{0.8\columnwidth}
        \centering
        \includegraphics[width=\textwidth]{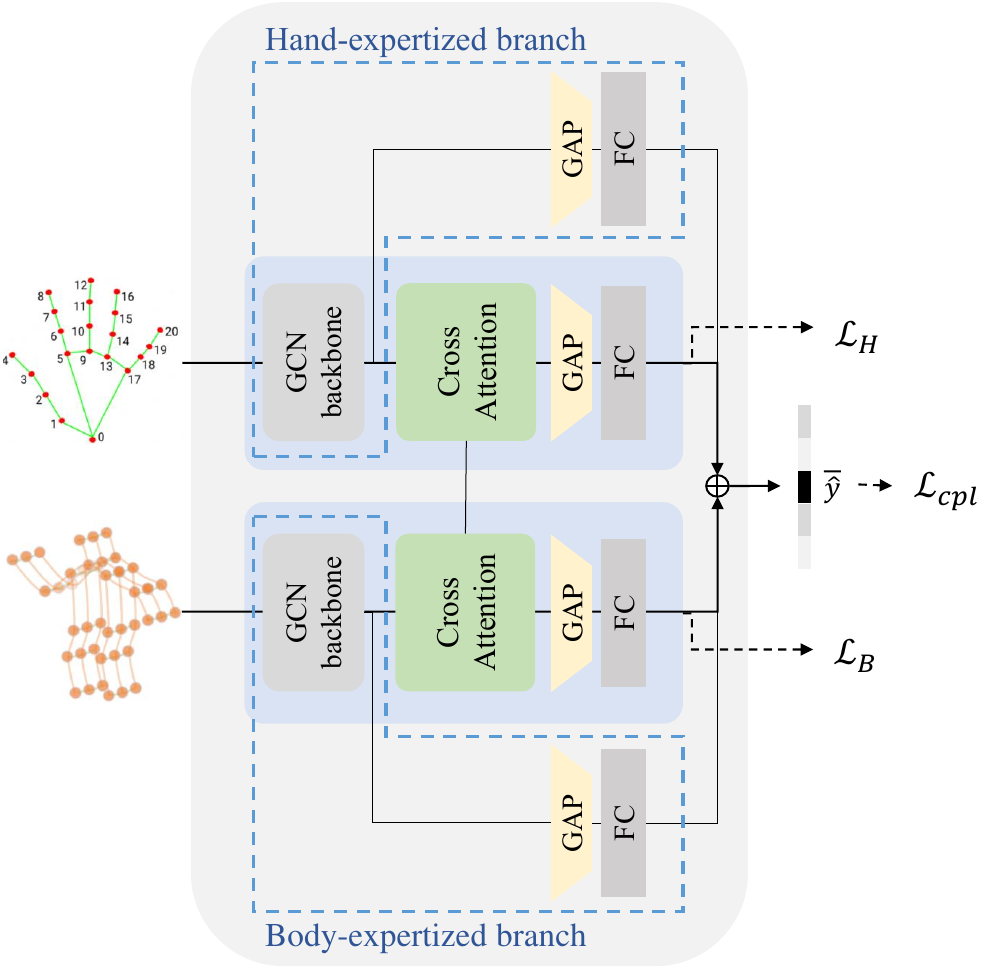}

        \label{fig:expert}
    \end{minipage}
    \begin{minipage}[t]{1.0\columnwidth}
        \centering
        \includegraphics[width=\textwidth]{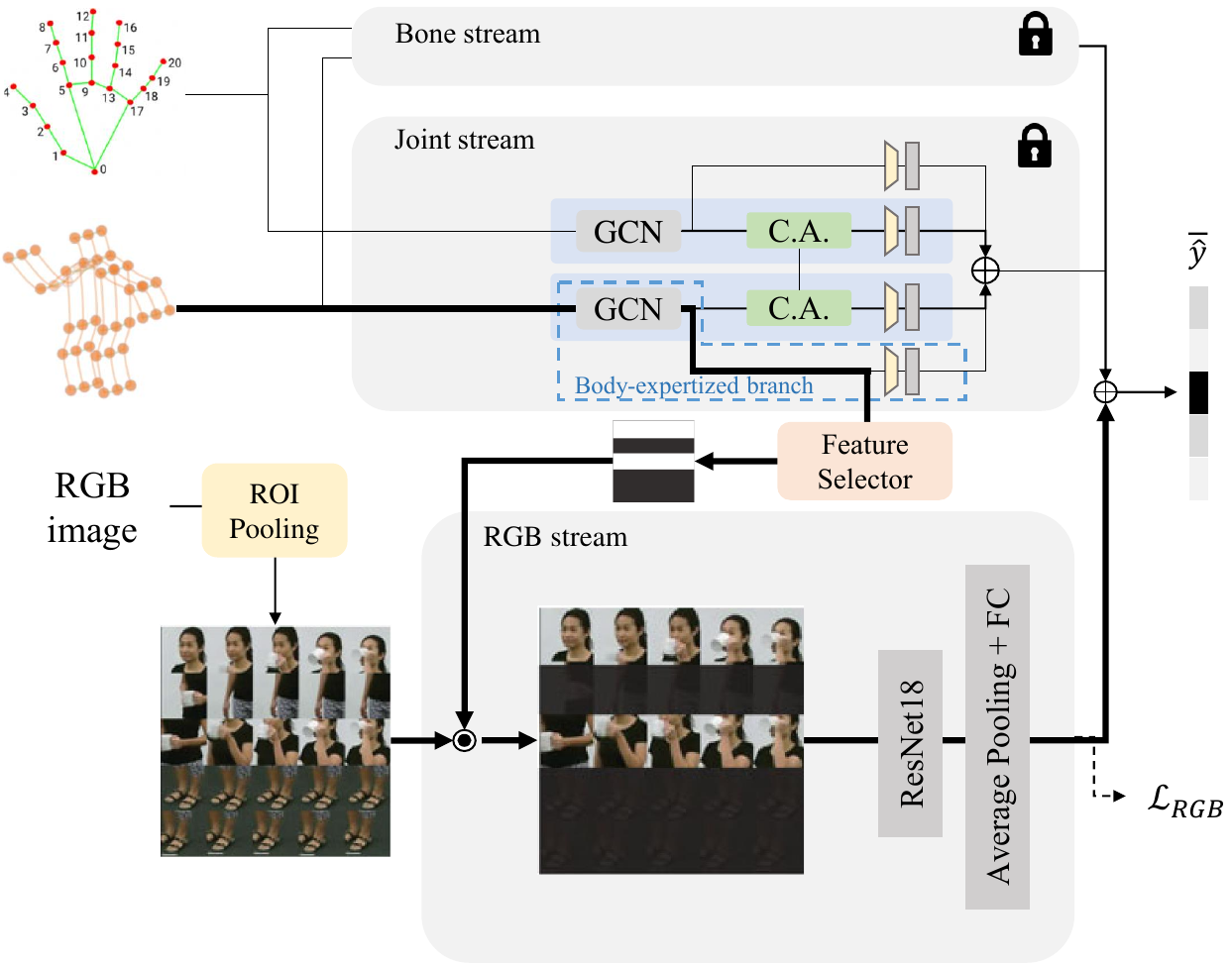} 

        \label{MMNet}
    \end{minipage}
    \label{fig:two_images}
    \caption{\textbf{Pipeline of our Expertized Branch Model(Left) and Multi-modal Architecture(Right).} 
        In Expertized Branch Model, the blue-colored boxes indicate \emph{interactive branches} and blue dashed boxes are expertized branches.
        We integrate the Expertized Branch Model for both the Joint and Bone streams,  and add an RGB stream with its own training path (bold lines). The RGB branch receives body-joint guidance from the body-expertized branch, focusing the visual feature extractor on relevant spatio-temporal regions.
        }
\end{figure*}

\subsection{Cross-Attention for Feature Interaction}
While preserving modality-specific features is essential, interaction between body and hand streams can provide valuable context. We incorporate a cross-attention mechanism enabling feature-level information exchange between modalities. This mechanism manifests through a pooling attention module.

\begin{equation}
    \tilde{\mathbf{F}_B}, \tilde{\mathbf{F}_H} = \mathrm{CrossAttn}(\mathbf{F}_B, \mathbf{F}_H)\\
\end{equation}

\noindent\textbf{Pooling Attention Module.}
\label{subsec:pooling_cross_attention}
To enchance computational efficiency while retaining rich spatio-temporal information, we introduce a \emph{Pooling Attention Module}. Instead of applying full cross-attention on high-dimensional feature maps, we perform structured pooling along different axes prior to attention computation.

We decompose feature maps into three different pooled representations, each stream partially pools along two axes (e.g. temporal $T$ and vertical $V$). Body and hand then exchange information via a fast attention-based interaction\cite{choromanski2020performeratt}, highlighting informative features from one modality in the other. Afterward, the remaining axis (e.g. instance $I$) is pooled, followed by a fully-connected (FC) layer to produce logits. Applying cross-attention between the partial pooling steps in dimensions $(T, V, I)$, we achieve a balanced trade-off between computational efficiency and rich feature integration.

Let the output of GCN backbone be 
$\mathbf{F} \in \mathbf{R}^{N \times C \times T \times I \times V}$,
where $N$ is the batch size, $C$ the feature channels, $T$ the temporal length, $I$ the number of instances(body or hand), and $V$ the spatial dimension. For $\alpha \in \{T, V, I\}, \beta, \gamma \in \{T,V,I\} \setminus \{\alpha\}$:
\begin{gather}
    \mathbf{F}_B^\alpha = \mathrm{GAP}_{(\beta, \gamma)}(\mathbf{F}_B),\quad\mathbf{F}_H^\alpha = \mathrm{GAP}_{(\beta, \gamma)}(\mathbf{F}_H)\\
    \tilde{\mathbf{F}_B^\alpha}, \tilde{\mathbf{F}_H^\alpha} = \mathrm{CrossAttn}(\mathbf{F}_B^\alpha, \mathbf{F}_H^\alpha)\\
    \hat{y_B} = \mathrm{FC}\bigl(\sum_{\alpha}\mathrm{GAP}_{(\alpha)}(\tilde{\mathbf{F}_B^\alpha})\bigr)\\
    \hat{y_H} = \mathrm{FC}\bigl(\sum_{\alpha}\mathrm{GAP}_{(\alpha)}(\tilde{\mathbf{F}_H^\alpha})\bigr)
\end{gather}

\subsection{Expertized Branch Model}
\label{subsec:expertized}
While cross-attention effectively captures feature-level interactions between modalities, it may unify their features in a way that risks obscuring subtle modality-specific cues. To mitigate this concern and leverage each modality’s distinctive strengths, we propose an \emph{Expertized Branch Model}, as depicted in Figure~\ref{fig:expert}.

In this model, the GCN outputs from body and hand each split into two parallel pathways: an \emph{Expertized branch} that maintains modality-specific features and an \emph{Interactive branch} that receives cross-attention signals from the complementary modality. The expertized branches(\(\mathbf{F}_1, \mathbf{F}_2\)) pass the raw GCN features directly to GAP and FC layer, thereby preserving each modality’s inherent characteristics. The interactive branches(\(\mathbf{F}_3, \mathbf{F}_4\)) incorporate cross-attention-based fusion before undergoing GAP and classification. We then sum the logits from all four branches to obtain the final prediction:
\begin{table*}[t]
    \centering
    \caption{Accuracy(\%), FLOPs, and Parameter Comparison with State-of-the-Art Methods on NTU RGB+D 60/120. “-" indicates the experimental results are not provided in the reference, and “*" for the result based on using public codes. Best results are highlighted as \textbf{first} and \underline{second}.}
    \label{tab:ntu_skel}
    \resizebox{0.8\linewidth}{!}{
    \begin{tabular}{lccccccccc}
    \toprule
    \multirow{2}{*}{Method} & 
    \multicolumn{2}{c}{NTURGB+D} &
    \multicolumn{2}{c}{NTURGB+D 120} &
    {N-UCLA} &
    \multirow{2}{*}{GFLOPs} & 
    \multirow{2}{*}{Params(M)} \\ \cmidrule(lr){2-3}\cmidrule(lr){4-5}\cmidrule(lr){6-6}
    
     & X-Sub & X-View & X-Sub & X-Set & X-View & \\
    \midrule
    CTR-GCN~\cite{chen2021ctrgcn} & 92.4 & 96.8 & 88.9 & 90.6 & 96.5 & 7.9 & 5.8 \\
    InfoGCN~\cite{chi2022infogcn} & 93.0 & 97.1 & 89.8 & 91.2 & 97.0 & 10.0* & 9.4\\
    PoseConv3D~\cite{duan2022poseconv} & 94.1 & 97.1 & 86.9 & 90.3 & - & 31.8 & 4.0 \\
    BlockGCN~\cite{zhou2024blockgcn} & 93.1 & 97.0 & 90.3 & 91.5 & 96.9 & 6.5 & \underline{5.2}\\
    DeGCN (baseline)~\cite{myung2024degcn} & 93.6 & 97.4 & 91.0 & 92.1 & 97.2 & 6.9 & 5.6\\
    3Mformer\cite{wang20233mformer} & 94.8 & 98.7 & 92.0 & 93.8 & \textbf{97.8} & 58.5 & 6.7 \\
    SkeleT~\cite{yang2024expressive} & \textbf{97.0} & \textbf{99.6} & \textbf{94.6} & \textbf{96.4}& \underline{97.6} & 9.6 & \underline{5.2}\\
    \midrule
    \textbf{Ours (BHaRNet-E, Expert-only)} & 96.1 & 98.7 & 94.0 & 94.9 & 94.6 & \textbf{3.3} & \textbf{2.8}\\
    \textbf{Ours (BHaRNet-E, All-ensemble)} & 96.2 & \underline{98.8} & \underline{94.3} & 95.0 & 94.6 & 5.44 & 5.5\\
    \textbf{Ours (BHaRNet-P)} & \underline{96.3} & \underline{98.8} & \underline{94.3} & \underline{95.2} & 95.3 & \underline{3.42} & 9.06\\
    \bottomrule
    \end{tabular}
    }
\end{table*}

\begin{table*}[t]
    \centering
    \caption{Multi-modal action recognition on NTU60/120, PKU-MMD. “*" denotes experimental results provided in the other paper.}
    \label{tab:ntu_multimodal}
    \resizebox{0.8\linewidth}{!}{
    \begin{tabular}{lcccccccc}
    \toprule
    \multirow{2}{*}{Method} &
    \multicolumn{2}{c}{NTU60} &
    \multicolumn{2}{c}{NTU120} &
    \multicolumn{2}{c}{PKU-MMD} &
    \multirow{2}{*}{GFLOPs} &
    \multirow{2}{*}{Params(M)} \\
    \cmidrule(lr){2-3}\cmidrule(lr){4-5}\cmidrule(lr){6-7}
     & X-Sub & X-View & X-Sub & X-Set & X-Sub & X-view& & \\
    \midrule
    MMNet (baseline)\cite{bruce2022mmnet} & 96.6 & 99.1 & 92.9 & 94.4 & \textbf{97.4} & \textbf{98.6} & 89.2 & 34.2 \\
    PoseConv3D\cite{duan2022poseconv}   & \textbf{97.0} & \textbf{99.6} & \textbf{95.3} & \textbf{96.4} & - & - & 41.8* & 31.6* \\
    \midrule
    \textbf{Ours(BHaRNet-M)} & 96.3 & 99.0 & 95.1 & 96.0 & 96.9 & 97.9 & \textbf{7.6} & \textbf{16.7} \\
    \bottomrule
    \end{tabular}}
\end{table*}
\begin{gather}
    \mathbf{F}_1 = \mathbf{F}_B,\quad\mathbf{F}_2 = \mathbf{F}_H\\
    \mathbf{F}_3 = \tilde{\mathbf{F}_B}, \quad\mathbf{F}_4 = \tilde{\mathbf{F}_H}\\
    \hat{y} = \sum\hat{y}_i,\quad \hat{y}_i = \mathrm{FC}\bigl(\mathrm{GAP}(\mathbf{F}_{i})\bigr)
\end{gather}

Here, $\mathbf{F}_1, \mathbf{F}_2, \mathbf{F}_3, \mathbf{F}_4$ indicates the branch of expert body, expert hand, interactive body, interactive hand, respectively. We also use a fast-attention-based cross-attention module to keep computational overhead manageable.  To explicitly encourage expertized branches to specialize further, we modify our individual loss formulation:
\begin{gather}
    \mathcal{L}_{B} = \mathrm{CE}\bigl(\hat{y_3},\,l\bigr),\quad
    \mathcal{L}_{H} = \mathrm{CE}\bigl(\hat{y_4},\,l\bigr)
\end{gather}

In addition, a complementary loss is applied across all branches, as defined in \Cref{subsec:loss}. By assigning individual losses only to the interactive branches, we avoid excessive merging of modality-specific features. The complementary loss further encourages each expert branch to focus on its inherent strengths, reducing the need to handle every class uniformly and thereby retaining expert-level cues from body or hand data.

\subsection{Applications}
The \emph{Expertized Branch Model} naturally extends to applications requiring lightweight or multi-modal setups due to its modular design.

\noindent\textbf{Light-weighted Inference.}
By selecting only the body- and hand-expertized branches for inference, we derive a simplified model that reduces computational load while retaining the specialized advantages learned from the shared backbone and complementary loss. Though it does not employ cross-attention, its focus on explicit body and hand experts can offer a pragmatic balance between accuracy and efficiency for scenarios where low-power constraints are critical.

\noindent\textbf{RGB Integration.}
Inspired by MMNet~\cite{bruce2022mmnet}, we incorporate RGB information by treating body-expert features as guidance for a separate video stream. This enables our framework to highlight relevant spatio-temporal regions in RGB frames, complementing the skeleton data. In practice, body skeleton data covers global motion, hand skeleton focuses on finer gestures, and RGB frames capture contextual cues such as objects or backgrounds. Together, these data sources enhance each other through cross-attention, resulting in a unified approach suited to a wide range of action recognition tasks.

\subsection{Implementations and Preprocessing method}
To address the absence of hand skeleton data in the datasets\cite{shahroudy2016ntu, liu2019ntu120, wang2014nucla, liu2017pku-mmd}, we employ Mediapipe~\cite{lugaresi2019mediapipe} for 2.5D hand skeleton extraction and hand-centric graph connections from RGB frames. We preprocessed the extracted data into a similar format with body skeleton. Because there is a limit of 2 people in the datasets, we choose 2 hands(left, right) from the most active person in the sample. By the simple selective method, we can filter the noise from the mixture of noisy features. 
The extracted hand data undergoes a hand-centric preprocessing pipeline with modified sort and interpolation process. Lastly, the hand data is adjusted to match the body data dimensions (by augmenting with dummy keypoints). This ensures compatibility with existing body skeleton data, enabling cohesive multi-modal learning.

\section{EXPERIMENTS}
\label{sec:experiments}
In this section, we demonstrate the effectiveness and efficiency of our proposed method through extensive evaluations on large-scale skeleton-based benchmarks and a multi-modal integration task. We also conduct ablation studies to examine the impact of key design choices such as complementary loss, pooling attention, and expertized branches.
We denote \textbf{BHaRNet-P} as our dual-stream network with pooling attention module. \textbf{BHaRNet-E} refers to the expertized branch model and \textbf{BHaRNet-M} denotes our multi-modal architecture.
\subsection{Expertimental Setup}
We adopt the open-source DeGCN framework\cite{myung2024degcn} as our skeleton baseline, which uses CTR-GCN\cite{chen2021ctrgcn} plus temporal shift modules(TSM). We changed two-stream architecture to single-stream for the better backbone to be used in our dual-stream network. Joint training is performed by fine-tuning pre-trained expertized models on the same dataset. 

For joint training, the expertized model is initialized from pre-trained weights on the same dataset and fine-tuned with our dual-stream architecture. For the inference, we ensemble our model for Joint and Bone for skeleton-based action recognition. And also for the multi-modal action recognition, the modality(Joint, Bone, RGB) are in ensemble.

%
%



\subsection{Skeleton-Based Action Recognition}
We first compare our skeleton-based model against existing methods on the NTU 60/120 and N-UCLA benchmarks. ~\Cref{tab:ntu_skel} compares our method with state-of-the-art methods in terms of accuracy, FLOPs, and parameters. SkeleT incorporates body, hand, and foot joints into a unified graph and achieves strong performance. 

Our method attains accuracy close to other SOTA approaches like 3Mformer and SkeleT, but uses significantly fewer FLOPs and parameters. For instance, on NTU 120 X-Sub, Ours(BHaRNet-P) reaches 94.3\% while keeping GFLOPs nearly half of BlockGCN’s. Although SkeleT yields marginally higher accuracy, our model saves \(\approx60\%\) of FLOPs, demonstrating a strong efficiency--accuracy trade-off. 

\begin{figure}[t]
\centering
\includegraphics[width=0.99\linewidth]{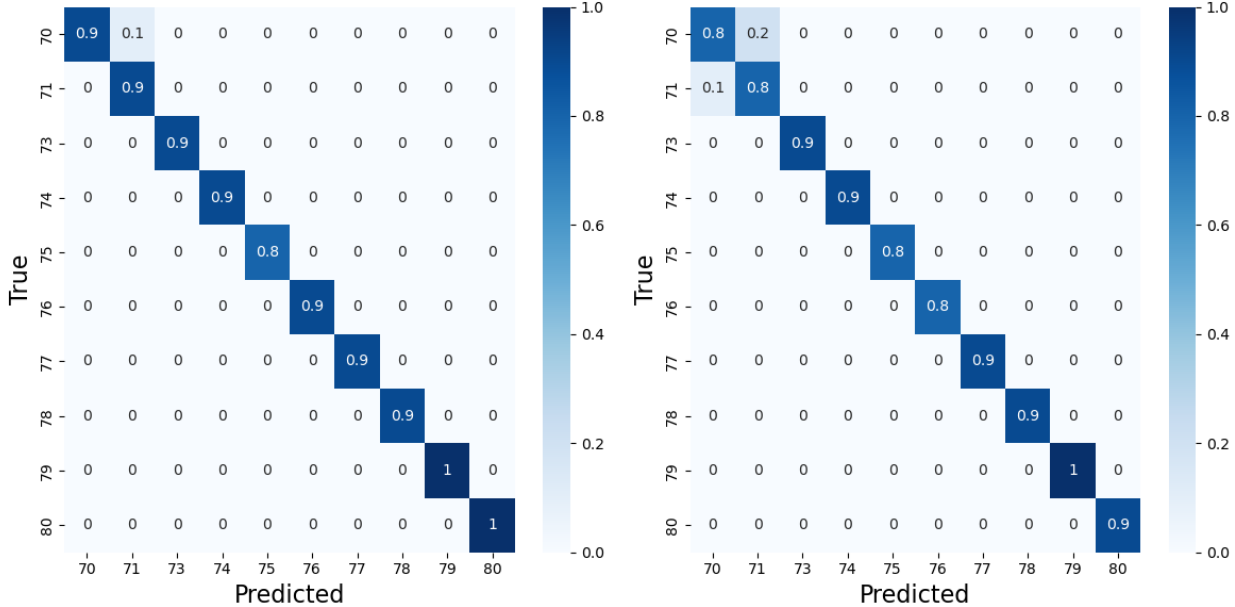}
\caption{Confusion matrix on hand-oriented actions(10 classes to visualize). Ours(BHaRNet-E) outperforms SkeleT by focusing on subtle finger articulations. Classes “Okay sign” and “Victory sign” mentioned in introduction are each index 70, 71 in the matrices.}
\label{fig:confusion_matrix}
\end{figure}

On top of that, a confusion matrix analysis(see \Cref{fig:confusion_matrix}) indicates that, for hand-specific actions like “OK sign” and “Victory sign”. BHaRNet-E exhibits notably fewer misclassifications than SkeleT. We attribute this to the hand branch’s specialized fine-motion features.
Although our models remain highly competitive on NTU 60/120, their performance on N-UCLA  does not reach the same level as SOTA methods. Further discussion of these aspects will be provided in \Cref{sec:limitation}.

%
%


\subsection{Multi-Modal Action Recognition}
We next test BHaRNet in a multi-modal(skeleton + RGB) setting. ~\Cref{tab:ntu_multimodal} summarizes the results on NTU 60/120 and PKU-MMD.
By taking body-expert features from BHaRNet-E as a guide to RGB streams, our multi-modal approach effectively localizes relevant spatio-temporal cues in the video. Although we do not surpass SOTA methods(MMNet, PoseConv3D), the gains are consistent overall, and we remain notably more parameter- and FLOP-efficient than typical CNN-based multi-modal pipelines. This suggests that our model(BHaRNet-M) can integrate seamlessly with existing multi-modal architectures while preserving efficiency and capturing subtle hand or body interactions that might be lost in monolithic designs.

\subsection{Ablation Studies}

%
\begin{table}[t]
    \centering
    \caption{Feature interaction on Joint NTU120 X-Sub. $L_{cpl}$ denotes for complemenatry loss. F.A. is shorten word of Fast Attention, and PAM is for Pooling Attention Module.}
    \label{tab:ablation_interaction}
    \resizebox{0.95\linewidth}{!}{
    \begin{tabular}{lcccc}
    \toprule
    \multirow{2}{*}{Method} & 
    \multicolumn{2}{c}{Accuracy(\%)} & 
    \multirow{2}{*}{GFLOPs} & 
    \multirow{2}{*}{Params(M)} \\
    \cmidrule(lr){2-3}
    & w/o $L_{cpl}$ & w/ $L_{cpl}$\\
    \midrule
    DeGCN(baseline, 2 stream) & 86.4 & - &  1.66 & 1.37 \\
    Body-expert(1 stream) & 86.3 & - & 0.83 & 0.69 \\
    Hand-expert(1 stream) & 70.1 & - & 0.83 & 0.69 \\
    \midrule
    Score fusion & 91.7 & 92.0 & 1.66 & 1.37 \\
    Cross-Attention(standard) & 92.5 & - & 1.88 & 2.46 \\
    Fast Attention~\cite{choromanski2020performeratt} & 92.7 & 92.7 & 2.72 & 4.33 \\
    \midrule
    \textbf{F.A.+ PAM} & 92.8 & \textbf{92.9} & \textbf{1.71} & 4.53\\
    \midrule
    \textbf{F.A.+Expertized branches} & 92.8 & \textbf{93.0} & 2.72 & \textbf{2.75} \\
    \bottomrule
\end{tabular}}
\end{table}

We conduct three ablation studies to reveal how complementary loss, cross-attention mechanisms, and expertized branches affect final performance.

\subsubsection{Joint Learning with Complementary Loss}
~\Cref{tab:ablation_interaction} demonstrates the effect of training the body and hand streams jointly using a complementary loss. Without this constraint, each stream can partially overlap in feature usage, leading to suboptimal synergy. Once complementary loss is enabled, the total ensemble accuracy increases(e.g. 91.7\% $\rightarrow$ 92.0\% on NTU120 X-Sub). 

\subsubsection{Feature Interaction Modules}
We compare diverse feature-fusion approaches in ~\Cref{tab:ablation_interaction}, including naive cross-attention, fast attention, and our proposed Pooling Attention Module. Fast attention improves accuracy over a simple cross-attention baseline, but pooling attention further boosts performance while controlling overall complexity. For instance, on NTU120 X-Sub, pooling attention attains 92.8\%($\rightarrow$ 92.9\% with complementary loss), outperforming the fast attention approach. Although the raw gain seems small, it is consistent across multiple seeds, and the additional overhead is modest. Hence, pooling attention can be seen as a balanced extension to fast attention that selectively focuses on the most relevant cross-modal features at each stage.

Pooling attention and expertized branches are built on the dual-stream baseline with fast attention. This synergy leads to consistent, if not dramatic, accuracy gains while retaining or even reducing model FLOPs. And also the performance enhancement in these modified architectures better supports for advantages of joint learning with complementary loss. Our main observation is that lightweight but well-structured fusion(pooling attention) can avoid overfitting or excessive overhead, supporting the design principle of cross-modal emphasis with experts.

\begin{table}[t]
    \centering
    \caption{Ablation study across backbones on Joint NTU120 X-Sub.}
    \label{tab:ablation_complementary}
    \resizebox{0.8\linewidth}{!}{
    \begin{tabular}{lcc}
    \toprule
    Model Variant & ST-GCN\cite{yan2018stgcn} & DeGCN \\
    \midrule
    Body & 83.2 & 86.3 \\
    Hand & 57.4 & 70.1 \\
    \midrule
    Simple Ensemble & 88.7 & 91.7\\
    Joint training, w/ $L_{cpl}$ & 89.0 & 92.0  \\
    \textbf{BHaRNet-P} & \textbf{89.6} & \textbf{92.9} \\
    \bottomrule
    \end{tabular}}
\end{table}
\begin{table}[ht]
    \centering

        \caption{Comparison of Body-Experts for body action guidance on Multi-modal NTU120 X-sub.}
        \label{tab:expert_vs_single_1}
        
        \resizebox{0.8\linewidth}{!}{%
            \begin{tabular}{lccc}
            \toprule
            Method & Pose & RGB & RGB+Pose \\
            \midrule
            Body GCN & 94.3  & 68.0 & 95.0 \\
            \textbf{Body-expertized branch}  & 94.3  & \textbf{71.0} & \textbf{95.1} \\
            \bottomrule
            \end{tabular}
        }
\end{table}
\begin{table}[ht]
    \centering
    \caption{Feature interaction within a pure Dual-stream Architecture on Joint NTU120 X-Sub.}
    \label{tab:expert_vs_single_2}
    
    \resizebox{0.8\linewidth}{!}{%
        \begin{tabular}{lc}
        \toprule
        Method & Accuracy(\%) \\
        \midrule
        Ensemble(simple GCN) & 91.7 \\
        Score fusion w/ $L_{cpl}$  & 92.0  \\
        \textbf{BHaRNet-E(Expert-only)} & \textbf{92.7}\\
        \bottomrule
        \end{tabular}
    }
\end{table}

\subsubsection{Ablation across Different Backbones}
We further validate our approach on two distinct GCN backbones---ST-GCN and DeGCN---and summarize the results in \Cref{tab:ablation_complementary}.

From these comparisons, we see that the improvement trend remains consistent across the two backbone architectures, suggesting that our proposed techniques---complementary loss and carefully designed cross-attention---are not tied to any particular GCN backbone. Even with ST-GCN, an earlier architecture known for lower representational capacity, we observe similar relative boosts from both joint training and pooling attention. This underscores the robustness of our design: by explicitly encouraging body--hand specialization and selectively unifying them at the right interaction points, BHaRNet consistently improves performance regardless of the underlying GCN backbone.


%

%

\subsubsection{Expertized Branch vs.\ Single Body GCN}

While previous ablations focused on complementary loss and feature interaction, we further investigate how an expertized body branch compares to a standard single body GCN in two scenarios: multi-stream fusion with RGB and a single-stream skeleton setting. \Cref{tab:expert_vs_single_1} presents results on NTU120(X-Sub) when combining pose with RGB input, while \Cref{tab:expert_vs_single_2} shows performance in a purely skeleton-based setup.

In \Cref{tab:expert_vs_single_1}, we compare a baseline body GCN model with our body-expertized branch, both integrated into an RGB pipeline. The body-expertized branch not only maintains strong pose-only performance(94.3\%) but also provides more informative body weights for the RGB stream, boosting the RGB-only accuracy. Consequently, the fused RGB+Pose performance also improves. Although the gains are incremental, they consistently appear across multiple runs, indicating that the body-expertized branch transfers clearer pose cues to the RGB network than a standard body GCN does.

\Cref{tab:expert_vs_single_2} investigates how the expertized branches perform in a pure dual-stream architecture(\Cref{fig:overview}). We compare a simple GCN ensemble, the same ensemble from joint training with complementary loss, and our full ensemble of expertized branches. The expert-only ensemble in BHaRNet-E outperforms both the naive score fusion and the ensemble with complementary loss. This result suggests that even when cross-attention is not present, distinct body and hand expert branches can better exploit their respective modality cues, leading to higher overall accuracy. By explicitly preserving separate experts, the network avoids merging body and hand features prematurely, yielding a more effective final ensemble.

Overall, these findings reinforce that body–hand expertization remains beneficial, whether in a dual-stream skeleton or a multi-modal fusion environment. Providing dedicated branches for each modality part(or region) not only enhances single-modality performance but also facilitates clearer weight guidance in multi-modal pipelines.

\section{LIMITATION}
\label{sec:limitation}
Although our model achieves competitive results on large-scale datasets, its accuracy on the smaller N-UCLA benchmark remains below certain specialized models. We find that the dataset’s limited size and relative scarcity of hand-centric classes reduce the benefits of our hand expert stream. In some cases, adding the under-trained hand branch even lowers the final ensemble accuracy compared to using the body stream alone. Future efforts could focus on improving the reliability of the hand stream, especially when hand-specific motions are sparse, to ensure that expert modeling remains advantageous despite limited data diversity.

\section{CONCLUSION}
We presented BHaRNet, a dual-stream skeleton-based action recognition framework that targets fine-grained hand motions and broader body dynamics through carefully designed cross-attention and expertized branches. Extensive evaluations on skeleton-based and multi-modal setups reveal that our method achieves near-SOTA accuracy at relatively low computational costs. Ablation studies confirm that (1) a complementary loss helps each stream specialize without redundancy, and (2) our two distinct feature-interaction approaches (Pooling Attention Module and Expertized Branches) improve performance and computational cost, albeit for different design reasons.

Although the improvements may appear incremental in absolute accuracy, they hold consistently across multiple backbones and tasks, indicating the broad applicability and efficiency advantages of our design. Future work may also explore a hybrid synergy of partial pooling with explicit specialization, or incorporate additional modalities (e.g., depth, inertial sensors) into the expertized architecture, continuing our pursuit of accurate yet lightweight multi-modal action recognition.

\section{ACKNOWLEDGEMENT}
We would like to thank the reviewers for their constructive feedback and suggestions. We also acknowledge the assistance of ChatGPT for editing support and grammar enhancements, which helped improve the clarity of the manuscript. We also extend our sincere gratitude to Hye-Yeon Kim for her valuable help and support throughout this work. This work was supported by NST grant (CRC 21011, MSIT), IITP grant (RS-2023-00228996, RS-2024-00459749, RS-2025-25443318, RS-2025-25441313, MSIT) and KOCCA grant (RS-2024-00442308, MCST).


\bibliography{root}
\bibliographystyle{IEEEtran}

\end{document}